\documentclass[letterpaper]{article} 
\usepackage{aaai2026}  
\usepackage{times}  
\usepackage{helvet}  
\usepackage{courier}  
\usepackage[hyphens]{url}  
\usepackage{graphicx} 
\usepackage{natbib}  
\usepackage{caption} 
\usepackage{algorithm}
\usepackage{algorithmic}

\usepackage{newfloat}
\usepackage{listings}
\DeclareCaptionStyle{ruled}{labelfont=normalfont,labelsep=colon,strut=off} 
\floatstyle{ruled}
\newfloat{listing}{tb}{lst}{}
\floatname{listing}{Listing}
\title{Jump-teaching: Combating Sample Selection Bias via Temporal Disagreement}
\author{
    Kangye Ji\textsuperscript{\rm 1, 2} \equalcontrib,
    Fei Cheng\textsuperscript{\rm 1} \equalcontrib \thanks{Corresponding Author.},
    Zeqing Wang\textsuperscript{\rm 1,3},
    Qichang Zhang\textsuperscript{\rm 4},
    Bohu Huang\textsuperscript{\rm 1}
}
\affiliations{

     \textsuperscript{\rm 1}School of Computer Science and Technology, \, Xidian University\\
    \textsuperscript{\rm 2}Tsinghua Shenzhen International Graduate School, \, Tsinghua University \\
    \textsuperscript{\rm 3}College of Design and Engineering, \, National University of Singapore \\
    \textsuperscript{\rm 4}Hangzhou Institute of Technology, \, Xidian University \\
    jky25@mails.tsinghua.edu.cn, chengfei@xidian.edu.cn, zeqing.wang@u.nus.edu, \\
    qichzhang@xidian.edu.cn, hbohu@xidian.edu.cn
}

\usepackage{bibentry}

\usepackage{amsmath}
\usepackage{amssymb}
\usepackage{url}            
\usepackage{booktabs}       
\usepackage{amsfonts}       
\usepackage{nicefrac}       
\usepackage{microtype}      
\usepackage{xcolor}         
\usepackage{adjustbox}
\usepackage{algorithm}
\usepackage{algorithmic}
\usepackage{amsmath}
\usepackage{subcaption} 
\usepackage{multirow}
\usepackage{array}
\usepackage{xr}
\usepackage{pifont}
\usepackage{esvect} 
\usepackage{tabularx}
\newcolumntype{Y}{>{\centering\arraybackslash}X}

\def\ie{\emph{i.e.,~}}
\def\eg{\emph{e.g.,~}}

\newtheorem{property}{Property}

\newcommand{\subfigref}[2]{Figure~\ref{#1}(\subref{#2})}

\renewcommand{\eqref}[1]{Equation~(\ref{#1})}

\usepackage{bm}
\def\1{\bm{1}}
\def\vx{{\bm{x}}}
\def\vz{{\bm{z}}}
\def\vp{{\bm{p}}}
\def\vzero{{\bm{0}}}

\begin{document}

\maketitle

\begin{abstract}
Sample selection is a straightforward technique to combat noisy labels, aiming to prevent mislabeled samples from degrading the robustness of neural networks. 
However, existing methods mitigate compounding selection bias either by leveraging dual-network disagreement or additional forward propagations, leading to multiplied training overhead.
To address this challenge, we introduce \textit{Jump-teaching}, an efficient sample selection framework for debiased model update and simplified selection criterion. Based on a key observation that a neural network exhibits significant disagreement across different training iterations, Jump-teaching proposes a jump-manner model update strategy to enable self-correction of selection bias by harnessing temporal disagreement, eliminating the need for multi-network or multi-round training. 
Furthermore, we employ a sample-wise selection criterion building on the intra variance of a decomposed single loss for a fine-grained selection without relying on batch-wise ranking or dataset-wise modeling. 
Extensive experiments demonstrate that Jump-teaching outperforms state-of-the-art counterparts while achieving a nearly overhead-free selection procedure, which boosts training speed by up to $4.47\times$ and reduces peak memory footprint by $54\%$.

\end{abstract}

\begin{links}
    \link{Code}{https://github.com/ky-ji/Jump-teaching}
    \link{Extended version}{https://arxiv.org/abs/2405.17137}
\end{links}

\section{Introduction}

Learning with Noisy Labels (LNL) is a critical challenge for real-world AI applications ~\citep{mahajan2018exploring,bakhshi2024balancing}. 
Typically, noisy labels stem from mistaken annotations of the dataset, such as in crowd-sourcing \citep{welinder2010multidimensional} and online query~\cite{blum2003onlinequery}. 
As accurate annotations of large datasets are a time-consuming endeavor, noisy labels become inevitable. Deep neural networks can easily overfit to noisy labels, which often leads to poor generalization performance~\cite{zhang2021understanding, han2020sigua}.



\begin{figure*}[t]
    \centering
    
    \begin{minipage}[b]{0.48\textwidth} 
        \centering
        \includegraphics[height=5cm, keepaspectratio]{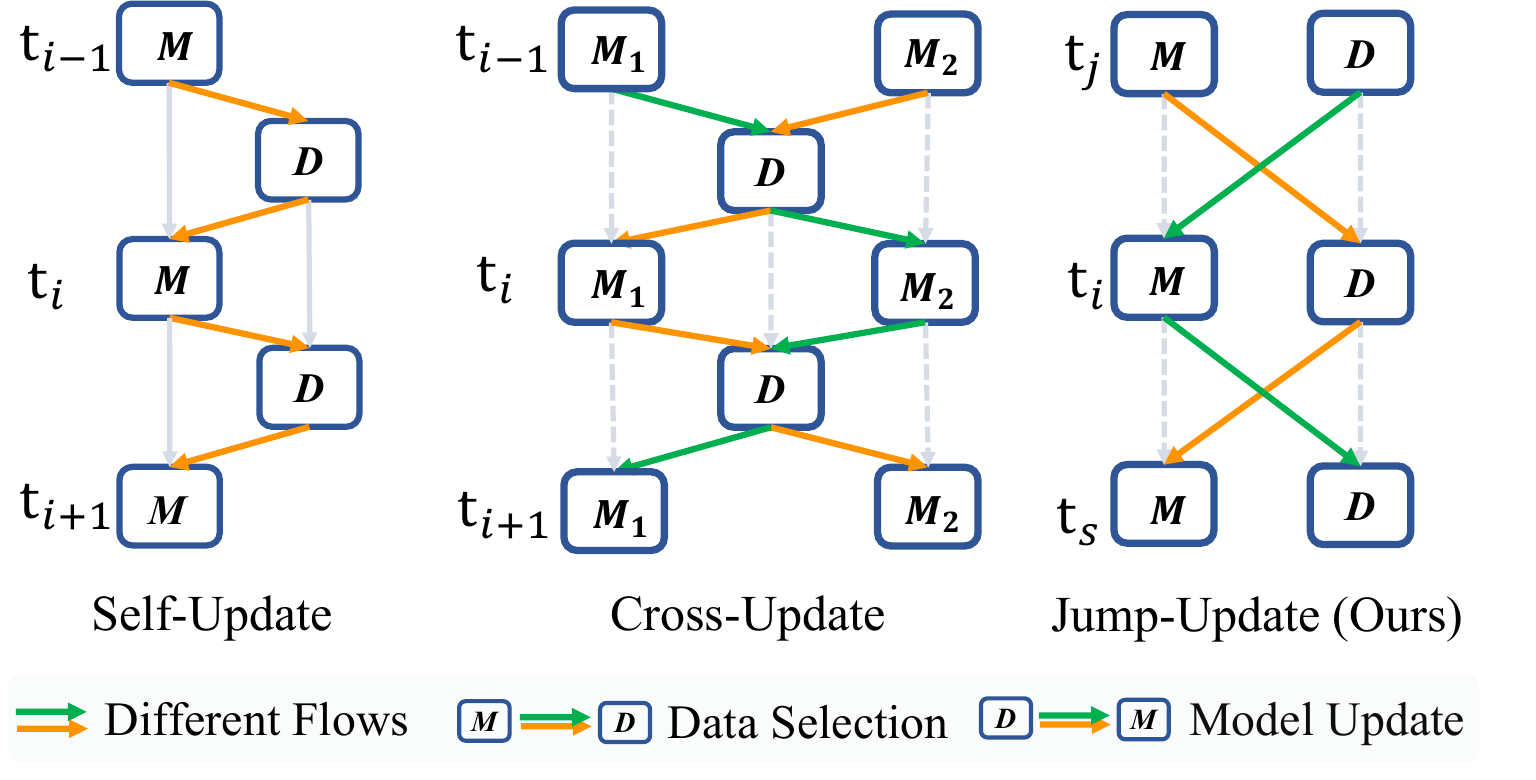}
        \subcaption{} 
        \label{fig:jump-update strategy}
    \end{minipage}
    \hfill 
    \begin{minipage}[b]{0.48\textwidth} 
        \centering
        \includegraphics[height=5cm, keepaspectratio]{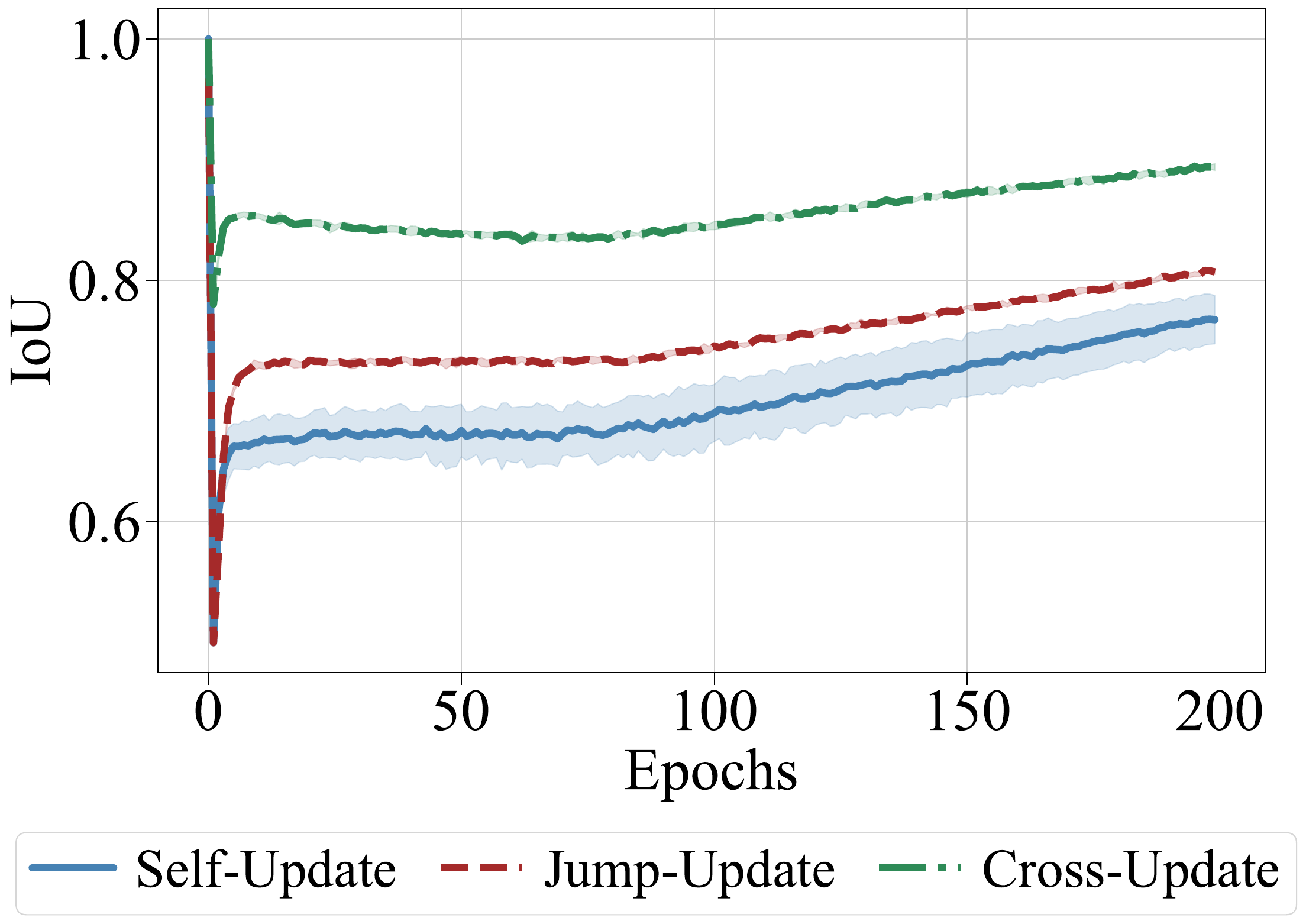}
        \subcaption{} 
        \label{fig: disagreement}
    \end{minipage}

    \vspace{-0.2cm} 
    
    \caption{(a)~The flow path of error under different update strategies. (b)~“Disagreement” of different strategies with symmetric noise ratio $\epsilon=0.5$ on CIFAR-10, quantified by the Intersection over Union (IoU) between two selections. In the cross-update strategy, disagreement arises between different networks, whereas in the other two strategies, it occurs across different iterations.}
    \label{Jump-update_all}
\end{figure*}

Despite sample selection representing a direct approach to combat label noise by selecting possibly clean samples from the corrupted dataset for training~\citep{song2019selfie,kim2021fine,wu2020topological,malach2017decoupling,xia2023combating, xiao2022promix}, selection bias has always been a fundamental obstacle for sample selection-based LNL methods. Selection bias refers to the tendency to favor noisy samples that have previously been selected and trained on~\cite{han2018co,yu2019does,wei2022selffiltering}. This bias inevitably arises from the exposure of the classifier to noise and causes the inclusion of noisy labels in the training data. When the neural network trains on these data, errors will accumulate and the bias will be compounded. 

Mitigating compounding selection bias demands costly training overheads. Explicit works typically utilize a dual-network training paradigm that employs the disagreement between networks in selection to correct the bias~\citep{malach2017decoupling,han2018co,tanaka2018joint,huang2023twin}, resulting in twice the standard training overhead. Implicit works avoid the bias from being frequently amplified by selecting samples from the entire dataset at once~\cite{bai2021me, yuan2023latestopping, yuan2025enhancing}, though these methods typically demand several times the computational resources.

In this paper, we introduce \textit{Jump-teaching}, an efficient framework to optimize the workflow of the sample selection-based LNL methods. Our key insight is that \textit{a neural network exhibits significant disagreement across different training iterations}, based on the observation in \subfigref{Jump-update_all}{fig: disagreement} that the intersection of the selected sets between different epochs of the network itself is much smaller than that between the two networks trained simultaneously. Motivated by this insight, we propose a jump-manner strategy termed \textit{Jump-update Strategy} for a debiased model update. As illustrated in \subfigref{Jump-update_all}{fig:jump-update strategy}, for each sample, our strategy generates its selection in iteration $t_j$ of the current epoch and updates the model based on the former selection conducted in iteration $t_i$ of the next epoch. In other words, the network in the previous epoch selects samples for the current network to train, which bridges the disagreement from a temporal perspective.

To achieve a training procedure without additional cost, we confine the selection process to operate solely at the mini-batch level. While the commonly adopted small-loss criterion \cite{han2018co} meets this constraint, its reliance on noisy rate prior and batch-ranking operations makes it impractical for standard training. Namely, we need a sample-wise selection criterion to fully unleash the potential of the jump-update strategy. However, one key issue is that a single loss value of a sample can't provide enough information to guide selection. To compensate for this lack of granularity, we decompose the loss for its intrinsic distribution, inspired by representation learning \cite{yang2015deep}. With richer information, we further propose the \textit{single-loss criterion}, based on the principle that \textit{a sample with a distorted intra-loss distribution is more likely to have a noisy label}. 
To operationalize this criterion, we design a lightweight plugin that contains an auxiliary head and a prepared codebook to transform the outputs and labels into semantic embeddings and hash encodings, respectively, within an auxiliary space. This plugin decomposes the single loss value into a vector with each dimension representing the sub-loss corresponding to a specific semantic feature. In this way, a noisy encoding yields larger sub-losses in dimensions affected by noise and smaller sub-losses in unaffected dimensions, exhibiting a distorted distribution in this vector. We then employ variance as a simple yet effective metric to assess this \textit{distorted} characteristic.

Benefiting from the jump-update strategy and the single-loss criterion, Jump-teaching updates the model with minimal bias and selects samples with finer granularity, achieving a new state-of-the-art across symmetric, asymmetric, pair-flip, real-world, and instance-dependent noise benchmarks. Moreover, it achieves this with an almost cost-free procedure, leading to $4.47 \times$ increase in training speed and $54 \%$ reduction in peak memory usage compared to previous methods. In summary, our main contributions are threefold:
\begin{itemize}
 \item [1.] Based on the observation that significant disagreement persists across training iterations, we propose the jump-update strategy, which enables a single network to effectively self-correct selection bias.
 \item [2.] We introduce Jump-teaching, an efficient sample selection framework, which employs the jump-update strategy for debiased model update and the single-loss criterion for sample-wise selection.
 \item [3.] Jump-teaching outperforms state-of-the-art methods on extensive noisy label benchmarks, particularly under extreme noise conditions. 
\end{itemize}

\section{Related Work}
\label{sec:Related}

\noindent \textbf{Learning with Noisy Labels}. 
Recent LNL methods can be divided into three categories: regularization, label correction, and sample selection. Regularization methods focus on crafting noise-robust loss functions \citep{ghosh2017robust, wang2019symmetric} and regularization techniques \citep{liu2020early, Zhang_2020_distill,cao2020heteroskedastic}, but they cannot fully avoid fitting to noisy labels during training, resulting in sub-optimal outcomes. Label correction, integrating closely with semi-supervised learning, aims to refine or recreate pseudo-labels \citep{ Han_2019_selflearn, sohn2020fixmatch, pham2021meta}. These methods make use of corrected noisy labels but require computational resources for the estimation of noise transition matrix \citep{goldberger2022adaptation} or the ensemble prediction \citep{lu2022selc}. 
The core idea of the sample selection approach is to filter out noisy samples to prevent the network from fitting them. The approach typically operates in an iterative workflow, selecting possibly clean samples through a certain criterion and then updating the model parameters based on those samples. 
We review key studies regarding selection bias and selection criteria in the next two paragraphs.

\noindent \textbf{Sample-selection Bias}.
\label{Previous Cross-Update Strategy} Sample selection inherently involves bias, leading to error accumulation. 
To mitigate the bias, existing methods either employ multiple networks or multiple rounds, leading to sub-optimal performance.
Decoupling \citep{malach2017decoupling} employs a teacher model to select clean samples to guide a student model, Co-teaching \citep{han2018co} simultaneously trains two networks on data selected by a peer network. Co-teaching+ \citep{yu2019does} maintains divergence by training only on samples where networks disagree while JoCoR \citep{Wei_2020_CVPR} employs regularization to remain divergent. Moreover, many methods integrate the co-training framework to achieve advanced performance \citep{li2020dividemix, liu2020early, liu2022robust, chen2023two}. However, these disagreement-based co-training methods double the computational overhead while relying on a weaker disagreement than the network itself, which motivates us to improve them. A rich body of research implicitly addresses bias by minimizing selection operations, which typically adopt a dataset-wise selection criterion. We will discuss them in the next paragraph.

\noindent \textbf{Sample-selection Criterion}.
Due to the lack of strong debiasing strategies, a dataset-wise selection criterion is preferred because it can not only improve the selection quality but also help prevent the repetitive amplification of bias.
One line of works~\citep{li2020dividemix,huang2023twin,kim2021fine} regards the sample selection as a binary classification problem, where the distribution of sample losses is modeled by techniques like Gaussian mixture models. Some other works \cite{wu2020topological,Iscen_2022_Neighbor} conduct nearest neighbor analysis on the entire dataset, regarding the outliers as noisy samples.
Several works~\citep{toneva2018empirical, wei2022selffiltering, yuan2023latestopping, yuan2025early} leverage sample-wise prediction differences between epochs for selection. They show evident redundancy in the additional forward propagations through the dataset before training and the repeated need to aggregate batches of data for noisy label estimation. Notably, although the observations of prediction differences share some commonality with our work, they have never been considered for debiasing.
Despite some methods \citep{han2018co,tanaka2018joint} relying on the small-loss criterion that ranks samples within each batch by their loss magnitude, prioritizing those with smaller losses, they necessitate prior knowledge about the noise rate, which is impractical in real-world scenarios.

\section{Methodology}
\label{sec:Jump}

We introduce Jump-teaching, an efficient sample selection framework. Specifically, it employs the \textit{Jump-update Strategy} 
to mitigate the selection bias and the \textit{Single-loss Criterion} for a sample-wise selection. We demonstrate them in Section~\ref{sub:jump_upate} and Section~\ref{sub:Semantic Loss Decompostion}, respectively.

\subsection{Jump-update Strategy}\label{sub:jump_upate}

Inspired by the observation in \subfigref{Jump-update_all}{fig: disagreement}, we propose the Jump-update Strategy, a jump-manner model update strategy designed to harness disagreement from ancestor iterations in a single network for bias correction. We provide a detailed strategy description, followed by an empirical quantitative analysis to investigate its debiasing mechanism and validate our insights through experimental analysis.

\noindent \textbf{Strategy Description.} To simplify our discussion, we dive into the procedure of sample selection and give some definitions. Sample selection is accomplished by the execution of model updates and sample selection with multiple iterations. In other words, model updates and sample selection are performed only once in every iteration. Previous strategies and ours are compared in \subfigref{Jump-update_all}{fig:jump-update strategy}. 

Suppose $t_i$ denotes the current $i$-th iteration, \textbf{ancestor iteration} refers to the former iteration $t_j, 0 \leq j \leq {i-2}$, excluding the previous 
$(i-1)$-th iteration, during the procedure of sample selection. Similarly, \textbf{descendant iteration} denotes the future iteration $t_s, i+1 \leq s \leq N_{\textrm{iterations}}$. $N_{\textrm{iterations}}$ is the total number of training iterations. The ancestor iteration and descendant iteration will not appear in the same epoch. In our paradigm, the current network in the $i$-th iteration is trained with clean samples selected only by the network from an ancestor iteration $t_j$. This behavior of sample selection exhibits a \textit{jump} manner. Concretely, we leverage a binary identifier to represent the outcome of the label judgment after sample selection. Thus, a binary identifier table $\mathcal{I}$ corresponds to the entire dataset. The jump-update strategy is divided into three steps: 
1) In the current iteration $t_i$, the network generates the new binary identifiers by the clean sample selection for the descendant iteration $t_s$. 
2) The network updates the parameters based on the clean data judged only by the old table. Before the update, the network inherits the weights from the previous iteration $t_{i-1}$. 
3) We update the table with identifiers cached in step 1.

The jump-update strategy harnesses the disagreement between different training iterations, enabling more effective bias correction due to a larger disagreement than the cross-update strategy. Suppose $S$ denotes the jump steps, it should be noted that $2 \leq S \leq N_{\textrm{iterations}}$. The setting of $S$ is illustrated in Section~\ref{sec:exp-jump-update}.

\begin{figure}[htbp]
    \centering
    \includegraphics[height=5.8cm, keepaspectratio]{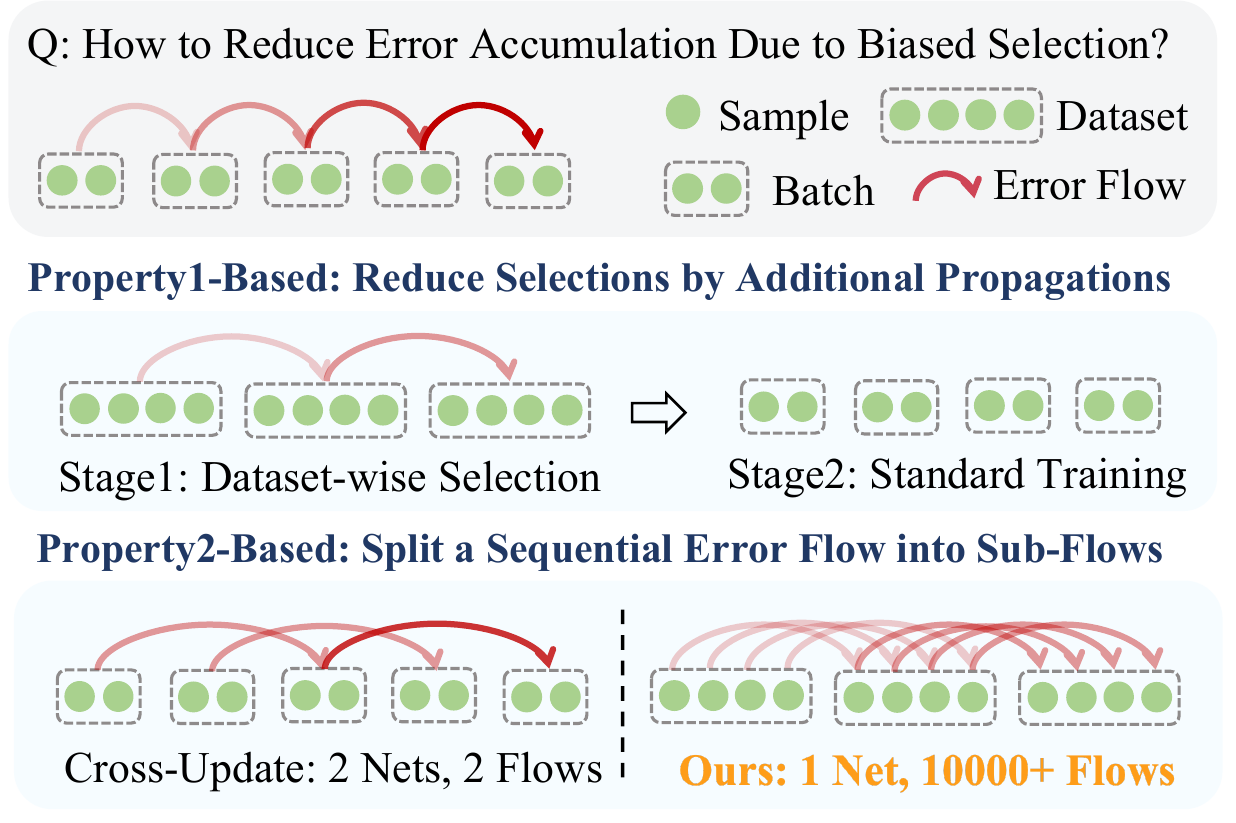} 
    \caption[Jump Update]{Error flows in Property 1-based and Property 2-based methods. The jump-update strategy reduces error accumulation without additional propagations or networks.}
    \label{fig:error flow}
\end{figure}

\noindent \textbf{Theoretical Analysis.} 
The bias introduces noise in the training data, leading to the accumulated error in updated parameters. Therefore, a core question in mitigating bias is: \textit{How can we reduce error accumulation caused by biased selection?} 
To answer this question, we utilize error flow to quantitatively represent the process of error accumulation, inspired by \cite{han2018co}. 
As shown in Fig. \ref{fig:error flow}, the graph of error flow presents a sequential form by iterations. 
Existing debiasing works weaken error flow through two properties: 1) reducing iterations or 2) splitting it into multiple sub-flows. 
Property 1-based methods require additional forward propagations as each iteration involves both selection operation and model update, while standard training procedures necessitate mini-batch updates. This constraint requires dataset-wise selection to minimize iterations. Previous property 2-based methods are built on the cross-update strategy, which split the flow into two sub-flows, as clearly demonstrated by the different colors in \subfigref{Jump-update_all}{fig:jump-update strategy}. Our jump-update strategy, however, limits the error accumulating orthogonally on different samples, extending the number of sub-flows to the dataset size. We formalize the aforementioned properties mathematically below before giving some notations.

\noindent \textbf{Notations.} Suppose that $N_A$ is the total number of error accumulations, $N_a$ is the number of error accumulations in an error sub-flow, and $N_f$ is the number of error sub-flows. Besides, the constant $e$ represents the number of training epochs and $n$ represents how many selections are made in one epoch. $D_A$ denotes the overall degree of accumulated error, while $d^k_a$ denotes the degree of accumulated error in the $k$-th error sub-flow. In the absence of a specific reference, we can also denote the degree of accumulated error in one error sub-flow by $d_a$.

\begin{property}
\label{proper: accumulated_equals}
The overall degree of accumulated error $D_A$ is proportional to the total number of error accumulation $N_A$, $D_A \propto N_A$. Under the hypothesis that the error flow is an uninterrupted model, the number of error accumulation $N_A$ equals the
total number of training iterations $N_{\textrm{iterations}}$, while $N_{\textrm{iterations}}$ equals $e\times n$. Therefore, $D_A \propto n$.
\end{property} 
The overall degree of accumulated error $D_A$ depends on $n$ stated in Property \ref{proper: accumulated_equals}. 
When the network selects data from a mini-batch, $D_A$ can be enormous because $n$ is equal to the number of mini-batches, \eg Co-teaching. If the network selects data from the entire dataset, $n$ can be reduced to $1$, thereby significantly reducing $D_A$, \eg TopoFilter  \citep{wu2020topological} and LateStopping \citep{yuan2023latestopping}. However, Property \ref{proper: accumulated_equals}-based methods are inefficient because an additional forward pass over the entire dataset before training is necessitated.

\begin{property}
\label{proper: jump_accumul_erro} The accumulated error could be reduced by splitting the error flow into multiple error sub-flows. The degree of accumulated error in the $k$-th error sub-flow $d^k_a$ is proportional to the number of error accumulations in each error sub-flow $N_a$, $d^k_a \propto N_a$. The number of error accumulations in each error flow $N_a$ equals $\frac{N_A}{N_f}$.

\end{property}

As stated in Property \ref{proper: accumulated_equals}, $D_A$ can be reduced by minimizing $n$, while $D_A$ can also be reduced by splitting into error sub-flows as illustrated in Property \ref{proper: jump_accumul_erro}. The first property relates to the sample selection mechanism, and the second property is associated with the different strategies of model updates: 1)~The self-update strategy follows a single error flow,  resulting in $d_a = D_A$, which leads to rapid error accumulation. 2)~The cross-update strategy has two error sub-flows and  $d_a=0.5D_A$, which mitigates the error accumulation to some extent. 3)~The jump-update strategy reduces $d_a$ to $0.5e$, which is a significantly minimal value.

\begin{figure}[htbp]
    \begin{subfigure}[b]{0.23\textwidth}
        \includegraphics[width=\textwidth]{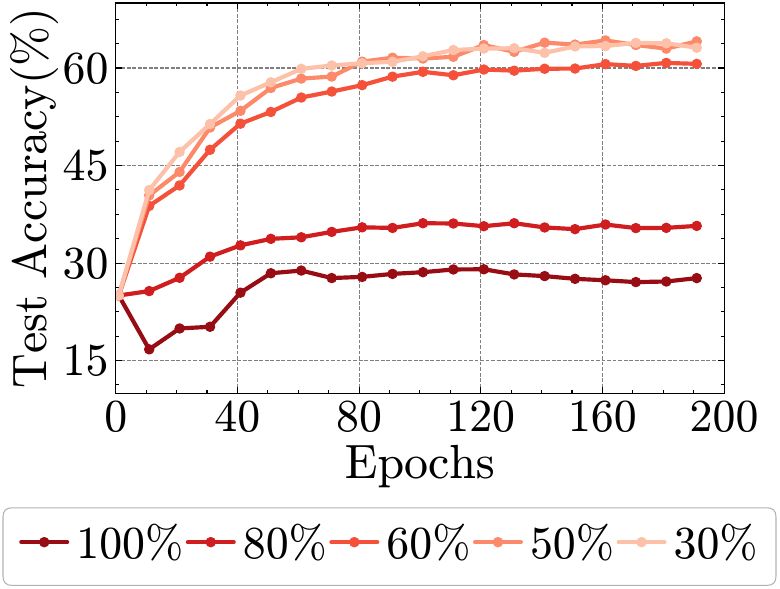}
        \caption{$D_A$ with different $n$}
        \label{fig:error accumulation}
    \end{subfigure}
    \begin{subfigure}[b]{0.23\textwidth}
        \includegraphics[width=\textwidth]{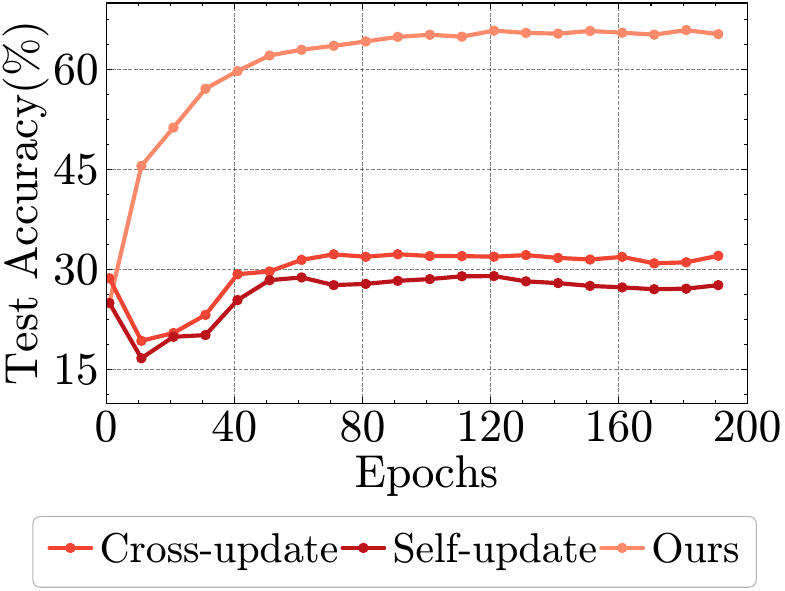}
        \caption{$D_A$ with different strategies}
        \label{fig:different strategies on error accumulation}
    \end{subfigure}
    \caption{Test accuracies on \textit{CIFAR-10} with Sym. $\epsilon=0.8$.}
    \label{fig:Obs_exp}
\end{figure}

\noindent \textbf{Experimental Analysis.} 
\label{sec: experimental analysis of error flow}
We verify Property \ref{proper: accumulated_equals} and Property \ref{proper: jump_accumul_erro} with toy examples, respectively. We employ the small-loss sample selection method from Co-teaching to establish the baselines for self-update and cross-update. Our experiments choose \textit{CIFAR-10} \citep{cifar10} dataset with the symmetric noise ratio $\epsilon = 80\%$. We utilize ResNet-18 \citep{ResNet} as the backbone and warm up it for one epoch before formal training. 

To verify Property \ref{proper: accumulated_equals}, we observe $D_A$ by testing the accuracy of the network with different values of $n$. 

To control $n$, We set $r$ as the proportion of sample selection that takes effects, $D_A$ equals $r\times n$ in this way. We set $r$ to $30\%$, $50\%$, $60\%$, $80\%$, and $100\%$. As shown in \subfigref{fig:Obs_exp}{fig:error accumulation}, rapidly accumulated errors lead to extreme deterioration of the model such as when $r=100\%$ and $r=80\%$, while slower accumulated ones achieve better performance, with a moderate $r=50\%$ yielding the best results. This is consistent with Property \ref{proper: accumulated_equals}. Our experiments not only verify Property \ref{proper: accumulated_equals}, but also reveal that simply setting $r$ offers an effective and cost-free alternative to mitigate the bias. 

To verify Property \ref{proper: jump_accumul_erro}, we observe $D_A$ by testing the accuracy of the neural network with three strategies: self-update, cross-update, and jump-update. As shown in \subfigref{fig:Obs_exp}{fig:different strategies on error accumulation}, cross-update slightly outperforms self-update, while jump-update is significantly more effective than both. Thus, splitting flows is an effective way to reduce error accumulation. 


\noindent \textbf{Remark.} 
The jump-update strategy is more effective and efficient than previous works. Compared to methods that implicitly leverage Property \ref{proper: accumulated_equals}, it can reduce $D_a$ to a small number \ie$0.5e$ at a constant cost. Compared to the cross-update strategy that leverages Property \ref{proper: jump_accumul_erro}, it can not only reduce the degree of accumulated error significantly more but also halve the training cost.

\subsection{Single-loss Criterion}
\label{sub:Semantic Loss Decompostion}

Due to the redundancy in the current sample selection criteria, we propose the single-loss criterion for sample-wise selection. Our design tackles two pivotal questions: 

\emph{Q1.\@ How to overcome the information insufficiency caused by relying solely on a single loss value?}

\emph{Q2. How to design a more effective selection criterion with the information enriched by addressing Q1?}

To address the first question, we propose analyzing the distribution within the loss across different semantic dimensions, drawing inspiration from prior work in representation learning \cite{yang2015deep, liu2016deep, yuan2020central}. In implementation, we design a lightweight plugin to decompose the single loss for a detailed distribution of sub-losses in different semantics. The plugin is structurally composed of a pre-prepared non-orthogonal codebook and an auxiliary head at the last layer of the network. It spans an auxiliary space where a loss can be decomposed semantically. Specifically, the codebook transforms the label into hash encodings while the head maps outputs into feature embedding. We will first detail the codebook and auxiliary head respectively, and then introduce the decomposed loss function by employing both.

\noindent \textbf{Codebook.} 
Inspired by \citet{yang2015deep}, we utilize the favorable properties of Hadamard matrices to construct mappings for category encoding. A $K$-bit Hadamard matrix can generate $2K$ codewords, each $K$ bits long, with a minimum Hamming distance of $\frac{K}{2}$. For $K$-bit hash encodings, we construct a $K \times K$ Hadamard matrix. From this, we select $c$ row vectors as category encodings, each with a Hamming distance of $\frac{K}{2}$. Noisy labels $\tilde{y}$ are mapped into hash encodings $\tilde{y}^\prime$ through this codebook. Given a classification task with $C$ classes, the mapping is formalized as: $$ H : \tilde{y}_i \in \{0,\cdots,C-1\} \rightarrow \tilde{y}_i^\prime \in \{-1,1\}^K.$$

\noindent \textbf{Auxiliary Head.} 
The auxiliary detection head is an additional three-layer MLP with a Tanh activation function. It shares the same feature extractor with the original classification head and maps the outputs $\displaystyle \vx$ of neural networks to K-bit feature embeddings $\displaystyle \vz$, as represented by the function: $$f: x_i \in \mathbb{R}^n \rightarrow z_i \in \mathbb{R}^K.$$ 
For an ideally clean sample, the distance $ \mathbf{d}(\mathbf{z}^{(t)}_i, \mathbf{y}_i) $ between the output  $\mathbf{z}_t $ of an ideally clean sample indexed $i$ at time $ t $ and its label $ \mathbf {y}  $ can be expressed as:
\begin{equation}
\lim_{t \to \infty} \mathbf{d}(\mathbf{z}^{(t)}_i, \mathbf{y}_{i}) = \mathbf{0},
\end{equation}
where the notation $\displaystyle \vzero$ denotes a vector of zeros, indicating that the distances across different dimensions uniformly converge towards zero as $t$ approaches infinity.

We employ binary cross-entropy \citep{ruby2020binary} to define the distance $\mathbf{d}$ between feature embeddings and hash encodings. It indicates how well the network captures the semantics of each dimension, formulated as:
\begin{equation}
\mathbf{d}(\mathbf{z}_i, \tilde{\mathbf{y}}_i') = -\left[\tilde{\mathbf{y}}_i' \odot \log(\mathbf{z}_i) + (1 - \tilde{\mathbf{y}}_i') \odot \log(1 - \mathbf{z}_i)\right],
\end{equation}
where the distance vector $\mathbf{d}(\mathbf{z}_i, \tilde{\mathbf{y}}_i') $ obtained through decomposition directly describes the distribution of sample loss in semantic space, which addresses the first question. 

Moreover, to supervise the detection head, we apply the arithmetic mean on the distance vector $\mathbf{d}(\mathbf{z}_i, \tilde{\mathbf{y}}_i') $. Therefore, the loss function can be articulated as: 
\begin{equation}
    \mathcal{L}_i^{BCE} = -\frac{1}{K} \sum_{j=1}^K \left[\tilde{y}_{ij}^\prime \log(z_{ij}) + (1-\tilde{y}_{ij}^\prime) \log(1-z_{ij})\right].
    \label{eq:BCELoss}
\end{equation}


\noindent \textbf{Selection Criterion.} 
To address the second question, we consider a sample with a distorted intra-loss distribution to have a noisy label. Our principle comes from memorization effects that neural networks prioritize learning simpler patterns from data \cite{zhang2021understanding}. 
The small-loss criterion leverages this effect: clean labels are learned first by the network, hence exhibiting smaller losses compared to noisy ones. However, the relative magnitude of losses is determined by comparison with other samples \eg rank sample losses with Top-k algorithm \citep{han2018co,jiang2018mentornet} or modeling loss distribution \citep{li2020dividemix,permuter2006study}. 
The single-loss criterion avoids such costly overheads by leveraging memorization effects in intra-loss distribution where the clean dimensions will be learned first and incur smaller sub-losses, while the noisy dimensions result in larger sub-losses. Therefore, a sample with a distorted intra-loss distribution is more likely to have noisy labels.
To identify whether a label contains noise, we use variance to characterize the distortion:
\begin{equation}
\begin{aligned}
\mathrm{Var}\bigl(\mathbf{d}(\mathbf{z}_i, \tilde{\mathbf{y}}_i')\bigr)
&= \frac{1}{K}
\Bigl(\mathbf{d}(\mathbf{z}_i, \tilde{\mathbf{y}}_i')
    - \mathcal{L}_i^{BCE}\mathbf{1} \Bigr)^T \\
&\quad \times
\Bigl(\mathbf{d}(\mathbf{z}_i, \tilde{\mathbf{y}}_i')
    - \mathcal{L}_i^{BCE}\mathbf{1} \Bigr).
\end{aligned}
\end{equation}

\begin{table}[t]
\centering
\fontsize{8pt}{9.6pt}\selectfont
\setlength{\tabcolsep}{0.8pt}  
\renewcommand{\arraystretch}{0.2}  
\begin{tabularx}{\linewidth}{l| Y Y | Y Y}
\toprule
\multicolumn{1}{c|}{Datasets} & \multicolumn{2}{c|}{CIFAR-10} & \multicolumn{2}{c}{CIFAR-100} \\
\midrule
\multicolumn{1}{c|}{Criterion} & Small-loss & Single-loss & Small-loss & Single-loss \\
\midrule
w/o Jump-update & 59.94 & 72.70 & 35.25 & 48.64 \\
w/  Jump-update & 69.26 & 84.93 & 36.15 & 51.11 \\
\bottomrule
\end{tabularx}
\caption{Ablation study on different criteria with symmetric noise $\epsilon=0.5$. We utilize ResNet18 as our backbone.}
\label{tab: criterion}
\end{table}

We set a unified threshold to distinguish clean samples from noisy samples. Since this threshold is expected to approach zero infinitely, we set it to $0.001$. The identifier $\mathcal{I}_{\textrm{detection}}$ is updated by the detection head as follows:
\begin{align}
\mathcal{I}_{\textrm{detection}} = 
\begin{cases}
    \text{True} & \text{if } \mathrm{Var}(\mathbf{d}(\mathbf{z}_i, \tilde{\mathbf{y}}_i')) \leq \tau, \\
    \text{False} & \text{otherwise}.
    \label{eq:C1}
\end{cases}
\end{align}

\begin{table*}[ht]
\centering
{
\adjustbox{width=\textwidth}{
\begin{tabular}{@{}l|cccc|ccccc@{}}
\toprule
\multirow{1}{*}{Dataset} & \multicolumn{4}{c|}{CIFAR-10} & \multicolumn{4}{c}{CIFAR-100} \\
\midrule
\multirow{1}{*}{Noise Type} & \multicolumn{3}{c}{Sym.} & \multicolumn{1}{c|}{Asym.} & \multicolumn{3}{c}{Sym.} & \multicolumn{1}{c}{Asym.} \\
\midrule
{Noise ratio} & {0.2} & {0.5} & {0.8} & {0.4} & {0.2} & {0.5} & {0.8} & {0.4} \\
\midrule
Standard & 84.6$\pm$0.1 & 62.4$\pm$0.3 & 27.3$\pm$0.3 & 75.9$\pm$0.4 & 56.1$\pm$0.1 & 33.6$\pm$0.2 & 8.2$\pm$0.1 & 40.1$\pm$0.2 \\
Decoupling('17)  & 86.4$\pm$0.1 & 72.9$\pm$0.2 & 48.4$\pm$0.6 & 83.3$\pm$0.2 & 53.3$\pm$0.1 & 28.0$\pm$0.1 & 7.9$\pm$0.1 & 39.9$\pm$0.4 \\
Co-teaching('18)  & 89.9$\pm$0.6 & 67.3$\pm$4.2 & 28.1$\pm$2.0 & 79.2$\pm$0.5 & 61.8$\pm$0.4 & 34.7$\pm$0.5 & 7.5$\pm$0.5 & 40.0$\pm$1.2 \\
Co-teaching+('19)  & 88.1$\pm$0.0 & 61.8$\pm$0.2 & 22.3$\pm$0.6 & 58.2$\pm$0.2 & 54.5$\pm$0.1 & 27.6$\pm$0.1 & 8.4$\pm$0.1 & 19.9$\pm$0.3 \\
PENCIL('19)  & 88.2$\pm$0.6 & 73.4$\pm$1.5 & 36.0$\pm$0.7 & 77.3$\pm$3.4 & 57.4$\pm$1.0 & 11.4$\pm$3.2 & 5.4$\pm$1.2 & 45.7$\pm$0.9 \\
Topofilter('20)  & 89.5$\pm$0.1 & 84.6$\pm$0.2 & 45.9$\pm$2.6 & 89.9$\pm$0.1 & 63.9$\pm$0.8 & 51.9$\pm$1.7 & 16.9$\pm$0.3 & 66.6$\pm$0.7 \\

FINE('21)  & 90.2$\pm$0.1 & 85.8$\pm$0.8 & 70.8$\pm$1.8 & 87.8$\pm$0.1 & 70.1$\pm$0.3 & 57.9$\pm$1.2 & 22.2$\pm$0.7 & 53.5$\pm$0.8 \\
SPRL('23) & 91.7$\pm$0.2 & 88.4$\pm$0.6 & 63.9$\pm$1.4 & 89.8$\pm$0.5 & 69.5$\pm$0.8 & 57.2$\pm$1.6 & 23.8$\pm$2.2 & 58.7$\pm$1.3 \\

LateStopping('23) & 91.1$\pm$0.1 & 66.0$\pm$0.7 & 32.3$\pm$1.9 &  82.7$\pm$0.3 & 69.4$\pm$0.2 & 45.4$\pm$0.3 & 12.8$\pm$0.2 & 55.4$\pm$0.3 \\

RML('24)& 92.2$\pm$0.1 & 88.3$\pm$0.3 & 35.3$\pm$1.7 & 79.8$\pm$4.0 & 67.2$\pm$0.2 & 62.0$\pm$0.5 & 15.7$\pm$0.8 & 64.5$\pm$0.6 \\
\noalign{\vskip 2pt} 
\hline
\noalign{\vskip 3pt} 
Jump-teaching & \textbf{94.8$\pm$0.1} & \textbf{92.2$\pm$0.1} & \textbf{84.1$\pm$1.1} & \textbf{90.7$\pm$0.3} & \textbf{72.7$\pm$0.5} & \textbf{67.1$\pm$0.2} & \textbf{40.0$\pm$1.1} & \textbf{68.4$\pm$0.7} \\
\bottomrule
\end{tabular}}
}
\caption{Average test accuracy(\%) on \textit{CIFAR-10} and \textit{CIFAR-100} with symmetric and asymmetric noise. The mean and standard deviation over three trials are presented. All methods employ PreActResNet-18 and train $200$ epochs. The best results are highlighted in bold.}
\label{tab: comparision_with_sota}
\end{table*}

\subsection{Training Pipeline}

In this section, we illustrate how the proposed plugin can assist in selection. 
During training, the detection head is trained by Eq. \ref{eq:BCELoss} while the classification head is trained by the cross-entropy loss. During inference, only the classification head takes effect.
However, the two heads converge at markedly different rates during training, causing the network to prematurely overfit noisy labels. Specifically, the cross-entropy loss used for the classification head is easier to optimize and converges much faster than the binary cross-entropy loss of the detection head. Consequently, errors accumulate before the detection head is sufficiently trained to perform effective selection.

To balance the convergence rate of the two heads, we employ temperature scaling \citep{guo2017calibration} to calibrate the label probabilities $\displaystyle \vp$  of the classification head. Thus, the soften softmax function is modified as follows:
\begin{equation}
\sigma_{\mathrm{softmax}}\left(p_{ij}\right)=\frac{\exp \left(p_{ij}/T\right) }{\sum_{j=1}\exp \left(p_{ij}/T\right)},
\end{equation}
where $T$ is a temperature scaling factor, controlling the convergence rates of the head.

To make use of the classifier head, we follow the widely accepted principle that, assuming the model is well-trained, predictions of clean samples should align with true labels \cite{wang2022promix}. Based on this principle, we apply a straightforward criterion, which further recovers the discarded clean labels. The clean table of the classifier head is evaluated as:
\begin{equation}
\mathcal{I}_{\textrm{classifier}} = (\hat{y}_i == \tilde{y}_i),
\label{eq:C2}
\end{equation}
where $\hat{y}_i = \arg \max_j p_i^j$ is the prediction label and  $p_i^j$ represents the probability of the $j$-th class for the $i$-th sample. $\tilde{y}_i$ denotes the label of the $i$-th sample. Finally, we can update the table by combining Eq. \ref{eq:C1} and Eq. \ref{eq:C2}:
\begin{equation}
\mathcal{I}^\prime  = \mathcal{I}_{\textrm{detection}} \lor \mathcal{I}_{\textrm{classifier}} .
\label{eq:C_plus}
\end{equation}

To validate the single-loss criterion, we compare it with the cross-update criterion under various settings. As shown in Table~\ref{tab: criterion}, the single-loss criterion achieves over a 10\% improvement compared to the cross-update strategy, highlighting its effectiveness.

\section{Experiments}
\label{sec:exp}

In this section, we first describe the experimental settings in Section~\ref{sec:exp-setup}. Then, we demonstrate the effectiveness of Jump-teaching, compared with the state-of-the-art in Section~\ref{sec: eq-jump-teaching}. As the jump-update strategy is the key component of our method, we thoroughly examine it in Section~\ref{sec:exp-jump-update}.

\subsection{Experimental Setup}
\label{sec:exp-setup}





\begin{table*}[t]
\centering
\adjustbox{width =\textwidth}{
\begin{tabular}{l l|ccccccccc|c}
\toprule
Dataset & Metric & APL & CDR & MentorNet & SIGUA & Decoupling & Co-teaching & Co-teaching+ & JoCoR & CoDis & Jump-teaching \\
\midrule
\textit{Clothing1M}& Top-1 & 54.46 & 66.59 & 67.25 & 65.37 & 67.65 & 67.94 & 63.83 & 69.06 & 71.60 & \textbf{71.93} \\
\midrule
\textit{Food-101}& Top-1 & 82.17 & 86.36 & 81.25 & 79.68 & 83.73 & 78.88 & 76.89 & 84.04 & 86.13 & \textbf{86.67} \\

\midrule

\multirow{2}{*}{\textit{WebVision}} & Top-1 & 62.30 & 62.84 & 63.00 & 57.38 & 62.54 & 63.58 & 61.18 & 63.33 & 63.80 & \textbf{75.60} \\
 & Top-5 & 84.02 & 84.11 & 81.40 & 78.92 & 84.74 & 85.20 & 83.30 & 85.06 & 85.54 & \textbf{90.41} \\

\midrule 

\multirow{2}{*}{\textit{ILSVRC12}} & Top-1 & 61.27 & 61.85 & 57.66 & 52.88 & 57.26 & 61.22 & 58.74 & 58.76 & 62.29 & \textbf{71.66} \\
 & Top-5 & 84.82 & 85.80 & 80.01 & 74.67 & 80.50 & 84.78 & 82.72 & 82.85 & 85.39 & \textbf{90.15} \\

\bottomrule

\end{tabular}}
\caption{Test accuracy (\%) on real-world noisy benchmarks.}
\label{tab: Clothing1M}
\end{table*}

\begin{table*}[!htbp]
\centering

{

\setlength\tabcolsep{11.6pt}

\begin{tabular}{@{}l|ccccc|cccccccc@{}}
\toprule
\multirow{1}{*}{Dataset} & \multicolumn{5}{c|}{CIFAR-10} & \multicolumn{4}{c}{CIFAR-100} \\

\midrule
\multirow{1}{*}{Noise type} & \multicolumn{4}{c}{Sym.} & \multicolumn{1}{c|}{Asym.} & \multicolumn{4}{c}{Sym.} \\

\midrule
\multirow{1}{*}{Methods/Noise ratio} & 0.2 & 0.5 & 0.8 & 0.9 & 0.4 & 0.2 & 0.5 & 0.8 & 0.9\\
\midrule
DivideMix
& 95.7 & 94.4 & 92.9 & 75.4 & \textbf{92.1} & 76.9 & \textbf{74.2} & 59.6 & 31.0 \\
\multirow{1}{*}{J-DivideMix}
 & \textbf{96.0} & \textbf{94.6} & \textbf{93.2} & \textbf{77.7} & 91.9 & \textbf{77.3} & \textbf{74.2} & \textbf{59.7} & \textbf{32.5} \\
\noalign{\vskip 2pt} 
\hline
\noalign{\vskip 3pt} 
\multirow{1}{*}{DivideMix with \(\theta_1\) test}
 & 95.0 & 93.7 & 92.4 & 74.2 & 91.4 & 74.8 & 72.1 & 57.6 & 29.2 \\
\multirow{1}{*}{J-DivideMix with \(\theta_1\) test}
 & \textbf{95.8} & \textbf{94.8} & \textbf{92.7} & \textbf{79.4} & \textbf{92.2} & \textbf{77.3} & \textbf{73.2} & \textbf{57.8} & \textbf{30.8} \\

\noalign{\vskip 2pt} 
\hline
\noalign{\vskip 3pt} 

\multirow{1}{*}{DivideMix w/o co-training }
 & \textbf{94.8} & 93.3 & 92.2 & 73.2 & 90.6 & 74.1 & 71.7 & \textbf{56.3} & 27.2 \\
\multirow{1}{*}{J-DivideMix w/o co-training}
& 94.7 & \textbf{94.0} & \textbf{92.3} & \textbf{79.3} & \textbf{91.3} & \textbf{74.2} & \textbf{71.7} & 55.3 & \textbf{27.6} \\
\bottomrule
\end{tabular}
}
\caption{Comparison of test accuracies(\%) for DivideMix and J-DivideMix on \textit{CIFAR-10} and \textit{CIFAR-100} with different noise. The mean value of the last ten epochs are reported.}
\label{tab: J-dividemix}
\end{table*}

\begin{table}[ht]
\centering
\setlength{\tabcolsep}{3 pt}{


\begin{tabular}{lccc}
\toprule

Method & Single Net & Throughput $\uparrow$ & Peak Mem $\downarrow$ \\
\midrule
Standard      & \ding{51} & 4.16            & 0.68 \\
Decoupling    & \ding{55}    & 1.72            & 1.50 \\
Co-teaching   & \ding{55}     & 1.81            & 1.53 \\
Co-teaching+  & \ding{55}     & 2.63            & 1.53 \\
PENCIL        & \ding{51} & 3.13            & 0.73 \\
Topofilter    & \ding{51} & 1.61            & 1.40 \\
FINE          & \ding{51} & 2.17            & 0.94 \\
SPRL          & \ding{51} & 2.94            & 0.77 \\
RML           & \ding{51} & 0.91            & 0.81 \\
Jump-teaching & \ding{51} & \textbf{4.07}   & \textbf{0.70} \\
\bottomrule
\end{tabular}
}

\caption{Comparison of training throughput (in kfps) and peak memory consumption (in GB) across methods.}

\label{tab: Efficency}
\end{table}

\noindent \textbf{Noisy Benchmark Datasets.} We verify the experiments on three benchmark datasets, including $\textit{CIFAR-10}$, $\textit{CIFAR-100}$ and $\textit{Clothing1M}$ \citep{Clothing}. These datasets are popular for evaluating noisy labels. Following the setup on  \citep{li2020dividemix,liu2020early}, we simulate two types of label noise: symmetric noise, where a certain proportion of labels are uniformly flipped across all classes, and asymmetric noise, where labels are flipped to specific classes, \eg $bird\rightarrow{airplane}$, $cat\leftrightarrow{dog}$. We assume $\epsilon$ denotes the noise ratio.

\noindent \textbf{Baselines.} To be more convincing, we compare the competitive methods of LNL. These methods are as follows: Standard, which is simply the standard deep network trained on noisy datasets, Decoupling \citep{malach2017decoupling}, Co-teaching \citep{han2018co}, Co-teaching+ \citep{yu2019does}, PENCIL \citep{yi2019probabilistic}, TopoFilter \citep{wu2020topological}, ELR \citep{liu2020early}, FINE \citep{kim2021fine}, SPRL \citep{shi2023self}, LateStopping~\citep{yuan2023latestopping}, RML~\citep{li2024rml}, APL \citep{APL}, CDR \citep{CDR}, MentorNet \citep{jiang2018mentornet}, SIGUA \citep{han2020sigua}, JoCoR \citep{Wei_2020_CVPR} and CoDis \citep{xia2023combating}.

\noindent \textbf{Implementation Details.} All experiments operate on a server equipped with an NVIDIA A800 GPU and PyTorch platform. In the following experiments, Jump-teaching \textit{almost} employs the same configuration. It trains the network for $200$ epochs by SGD with a momentum of $0.9$, a weight decay of $1e-3$, and a batch size of $128$. The initial learning rate is set to $0.2$, and a cosine annealing scheduler finally decreases the rate to $5e-4$. The warm-up strategy is utilized by Jump-teaching, and the warm-up period is $30$ epochs. Exceptionally, we set the weight decay as $5e-4$ to facilitate learning on fewer available samples when the noise ratio $\epsilon$ equals $ 50\%$ and $80\%$ in \textit{CIFAR-100}, respectively. 
The single network of Jump-teaching employs three types of backbone networks. The baseline methods fully follow the experimental setup in the literature~\citep{han2018co,li2020dividemix}.

\subsection{Comparison with the State-of-the-Arts}
\label{sec: eq-jump-teaching}

\noindent \textbf{Synthetic Noisy Benchmark.} 
To verify the effectiveness of Jump-teaching, we compare our proposed method with a range of representative sample selection approaches. As shown in Table~\ref{tab: comparision_with_sota}, Jump-teaching demonstrates superior performance with all noise settings, confirmed as statistically significant ($p < 0.05$) by a Wilcoxon signed-rank test. Its superiority is particularly evident under extreme noise, where it achieves accuracy improvements of $13.3\%$ and $16.2\%$ on two datasets with a symmetric noise ratio of $\epsilon = 0.8$.

\noindent \textbf{Real-World Noisy Benchmark.} 
We operate real-world experiments including the \textit{Clothing1M}, \textit{Food-101}, \textit{WebVision}, and \textit{ImageNet ILSVRC12} datasets. 
Following the setup in \citet{xia2023combating}, we employ ResNet-50 pre-trained on \textit{ImageNet} \citep{Imagenet} dataset for \textit{Clothing1M} and \textit{Food-101}, and InceptionV2 for \textit{Web-Vision}, and \textit{ILSVRC12}. As shown in Table \ref{tab: Clothing1M}, the robustness of Jump-teaching for real-world noisy labels is favorable when compared to other methods.

\noindent \textbf{Efficiency Analysis.} 
We compare the efficiency of Jump-teaching with the above representative methods. The efficiency of these methods is evaluated by throughput and peak memory usage. 
Throughput, expressed in \textit{thousands of frames per second} (kfps), measures the training speed. It is calculated as the total number of training samples divided by the average time per epoch. Peak memory usage refers to the maximum amount of memory consumed during training. We evaluate these methods with symmetric noise ratio $\epsilon = 0.5$ in Table \ref{tab: Efficency}, while the accuracy of these methods is illustrated in Table \ref{tab: comparision_with_sota}. As shown in Table \ref{tab: comparision_with_sota} and Table \ref{tab: Efficency}, our method achieves almost up to $4.47\times$ training speedup, $54\%$ peak memory footprint reduction, and superior robustness overall.

\subsection{Experiments on Jump-update Strategy}
\label{sec:exp-jump-update}


The jump-update strategy is not only suitable for Jump-teaching but also easily integrated into other methods in LNL. DivideMix not only uses two networks to exchange error flow but also employs them to create pseudo-labels collaboratively. Consequently, we establish three baselines following \citet{li2020dividemix}: DivideMix, DivideMix with $\theta_1$ test, and DivideMix w/o co-training. DivideMix with $\theta_1$ test utilizes a single network for sample selection, while DivideMix w/o co-training retains a single sample and employs self-update for updating. In each case, we replace the original update strategy with the jump-update strategy. As Table \ref{tab: J-dividemix} shows, the jump-update strategy significantly improves the accuracy of all baselines, particularly under extreme noise ratios, indicating its effectiveness in combating selection bias.

\section{Conclusion}
\label{sec:conclusion}

In this paper, we propose Jump-teaching, an efficient sample selection framework to combat label noise. Our work explores the sample-selection bias mechanism and effectively mitigates the bias in a single network by leveraging temporal disagreement, especially under extreme noise. Moreover, with the proposed single-loss criterion, Jump-teaching achieves sample-wise sample selection, fully unleashing the power of the jump-update strategy. Extensive experimental validations confirm that Jump-teaching achieves SOTA performance in both robustness and efficiency.

\section*{Acknowledgments}
We would like to thank Suqin Yuan for the insightful discussions and constructive suggestions. The work was supported by the Fundamental Research Funds for the Central Universities (No. XJSJ25005), Ministry of Education Top-notch Student Training Program in Basic Disciplines 2.0 Research Topics (No. 20252012),  the Natural Science Basis Research Plan in Shaanxi Province of China (No. 2025JC-JCQN-089) and the Outstanding Youth Science Foundation of Shaanxi Province under Grant 2025JC-JCQN-083. We also acknowledge the support from the National Experimental Teaching Demonstration Center for Computer Network and Information Security affiliated with Xidian University.

\bibliography{aaai2026}

\end{document}